\documentclass[acmsmall,manuscript,screen,nonacm]{acmart}

\usepackage{relsize}
\usepackage{bigstrut}
\usepackage{multirow} 
\usepackage{multicol}
\usepackage{cleveref}
\usepackage{booktabs}
\usepackage[acronym]{glossaries}

\newcommand{\ACRONYM}{\textsc{No-IDLE}}

\begin{document}

\title{A look under the hood of the Interactive Deep Learning Enterprise (No-IDLE)}


\author{Daniel Sonntag}
\orcid{0000-0002-8857-8709}
\email{daniel.sonntag@dfki.de}
\author{Michael Barz}
\email{michael.barz@dfki.de}
\orcid{0000-0001-6730-2466}
\author{Thiago Gouvêa}
\email{thiago.gouvea@dfki.de}
\orcid{0000-0002-0727-5838}
\affiliation{
  \institution{German Research Center for Artificial Intelligence (DFKI)}
  \city{Oldenburg \& Saarbrücken}
  \country{Germany}
}


\begin{abstract}
This DFKI technical report presents the anatomy of the No-IDLE prototype system (funded by the German Federal Ministry of Education and Research) that provides not only basic and fundamental research in interactive machine learning, but also reveals deeper insights into users’ behaviours, needs, and goals.  Machine learning and deep learning should become accessible to millions of end users. No-IDLE's goals and scienfific challenges centre around the desire to increase the reach of interactive deep learning solutions for non-experts in machine learning. One of the key innovations described in this technical report is a methodology for interactive machine learning combined with multimodal interaction which will become central when we start interacting with semi-intelligent machines in the upcoming area of neural networks and large language models.   

\end{abstract}



\maketitle

\newacronym{dl}{DL}{deep learning}
\newacronym{dnn}{DNN}{deep neural network}
\newacronym{fsl}{FSL}{few shot learning}
\newacronym{hci}{HCI}{human-computer interaction}
\newacronym{idl}{IDL}{Interactive Deep Learning}
\newacronym{iml}{IML}{Interactive Machine Learning}
\newacronym{lime}{LIME}{local interpretable model-agnostic explanations}
\newacronym{ml}{ML}{machine learning}
\newacronym{mmi}{MMI}{multimodal-multisensor interfaces}
\newacronym{nli}{NLI}{natural language inference}
\newacronym{nlp}{NLP}{natural language processing}
\newacronym{qa}{QA}{question answering}
\newacronym{vqa}{VQA}{Visual Question Answering}
\newacronym{vr}{VR}{virtual reality}
\newacronym{xai}{XAI}{explainable AI}

\section{Introduction}

In recent years, machines have surpassed humans in the performance of specific and narrow tasks such as some aspects of image recognition or decision making along clinical pathways in the medical domain (weak AI). Although it is very unlikely that machines will exhibit broadly-applicable intelligence comparable to or exceeding that of humans in the next 30 years (strong AI), it is to be expected that machines will reach and exceed human performance on more and more applied tasks.
To develop the positive aspects of AI, manage its risks and challenges, and ensure that everyone has the opportunity to help in building an AI-enhanced society and to participate in its benefits, in this project, human intelligence and \gls{ml} take the centre stage: \emph{\gls{iml} is the design and implementation of algorithms and intelligent user interface frameworks that facilitate \gls{ml} with the help of human interaction.} 

Our focus is to improve the interaction between humans and machines, by leveraging state-of-the-art \gls{hci} approaches, as well as solutions that involve state-of-the-art \gls{ml} techniques.
In this project, we focus on \gls{idl}: \gls{dl} approaches for \gls{iml}.
We want computers to learn from humans by interacting with them in natural language for example and by observing them. 
Our goal in \ACRONYM{} \footnote{https://www.dfki.de/en/web/research/projects-and-publications/project/no-idle} is to improve the interaction between humans and machines to update \gls{dl} models, by leveraging both state-of-the-art human-computer-interaction and \gls{dl} approaches.
Basic and fundamental research in this corridor project should also reveal deeper insights into users' behaviours, needs, and goals. 
Machine learning and \gls{dl} should become accessible to millions of end users, and be functionally more advanced than current recommender systems in online shops that provide suggestions for items that are most pertinent to a particular user.
Explicit (ontological) knowledge representation and reasoning capabilities are however not part of this focused project, but a follow-up project would highly benefit from them.
In addition, we emphasise the role of multimodal interaction and 
mixed-initiative interaction. While focusing on \gls{idl} in this corridor project, we pose the development of a methodology for \gls{idl} as a challenge problem. A methodology for \gls{idl} will become central when we start interacting more with semi-intelligent machines. As a layer used to represent the interactions, opinions and feedback, it is critical that IML is well understood and defined.
Also, there has been recent and relatively rapid success of AI and \gls{ml} solutions that arise from neural network architectures. But neural networks lack the interpretability and transparency needed to understand the underlying decision process and learned representations. Making sense of why a particular model misclassifies test data instances or behaves poorly at times is a challenging task for model developers and is an important problem to address \cite{hohman2018visual}. A related argumentation is that despite their huge successes, largely in problems which can be cast as classification problems, the effectiveness of neural networks is still limited by their un-debuggability, and their inability to “explain” their decisions in a human understandable and reconstructable way \cite{10.1007/978-3-319-99740-7_21}. 

In \ACRONYM{}, we explore the relationship between \gls{dl}, \gls{hci}, and \gls{xai}. For example, by approaching the problem from the \gls{hci} perspective, recent work has shown the benefits of visualising complex data in \gls{vr}, e.g., in data visualisation \cite{7004282}, and big data analytics \cite{7322473}. In one \gls{hci} subtask in \ACRONYM{} for example, we extend an interactive image clustering method in \gls{vr} \cite{prange2021demonstrator}, where the user can explore and then fine-tune the underlying \gls{dl} model through intuitive hand gestures. While \gls{hci} constitutes a key approach, we will attack the \gls{iml} problem from multiple angles. Informed by emerging directions in both research and commercialisation of \gls{iml} systems \cite{zacharias2018survey, oviatt2019handbook}, we will deploy our expertise in \gls{mmi} and \gls{nlp}, while also tapping on the broader interdisciplinary community, to deliver on the mission to improve interaction between humans and machines.
Past application projects of DFKI's IML group include deep active learning such as described in \cite{shui2020deep}, explanatory interactive image captioning \cite{biswas2020towards}, \gls{idl} systems for melanoma detection \cite{sonntag2020skincare} and wildlife monitoring \cite{DBLP:conf/ijcai/GouveaKTLSCALST23}, toolkits for building multimodal systems and applications \cite{oviatt2019handbook,barz_multisensor-pipeline_2021}, and interactions with \gls{ml} systems as domain-specific explanations~\cite{hartmann2021interaction}. In \ACRONYM{}, we bring these approaches, technologies and our experience together to apply them to a special use case, namely interactive photo book creation, to test and evaluate the basic and fundamental research in this corridor project. The proposed project builds upon this broad experience and research results of the IML group in the areas of human-computer interaction (HCI), machine learning (ML), multimodal human‐computer interaction (MMI), and natural language processing (NLP).

In a nutshell, in \ACRONYM{} we explore \gls{idl} from four different perspectives (HCI, ML, NLP, MMI). \ACRONYM{} is a basic research project to advance our understanding of \gls{iml}.
We expect practical contributions to be made while bringing the four working groups of IML closer together to work on a specific application around IML for photo book creation and the exploitation of the findings in ongoing DFKI consortial and industrial projects.

\section{Use case: Interactive Photo Book Creation}
\label{sec:usecase}
The research questions raised in \ACRONYM{} will be investigated in the context of a specific use case: the interactive creation of a photo book. Consider the following scenario:

Family Smith (a family of four) takes many photos from all kinds of events and occasions and regularly likes to create personal photo books and calendars for themselves and as gifts for family members and friends. Selecting the right photos, arranging them and writing captions is fun but very time consuming, and while they appreciate it as a means of their personal expression and creativity, they would like to speed up the process, especially with respect to the more tedious parts like selecting among similar photos or finding a basic arrangement. At the same time, they would like to maintain control and a personal connection to the results. Each family member has their own personal taste: some are more inclined to funny situations and photos of people, other prefer scenic views and interesting lighting and their personal style of arrangement, some like to put the photos simply side by side, others like to make use of interesting frames, clip art and creative arrangements. In addition, the goal and target audience influence their choices. For instance, they like to create diary type photo books of their travels for their own archive but like to tell image stories of the same trip for showing them or gifting them to others. When they create books or calendars for special holidays or birthday gifts, they typically select photos that somehow match the occasion but that also contain the gifted person if possible.

\begin{figure}
    \centering
    \includegraphics[width=\textwidth,trim={0 10.5cm 4.8cm 0},clip]{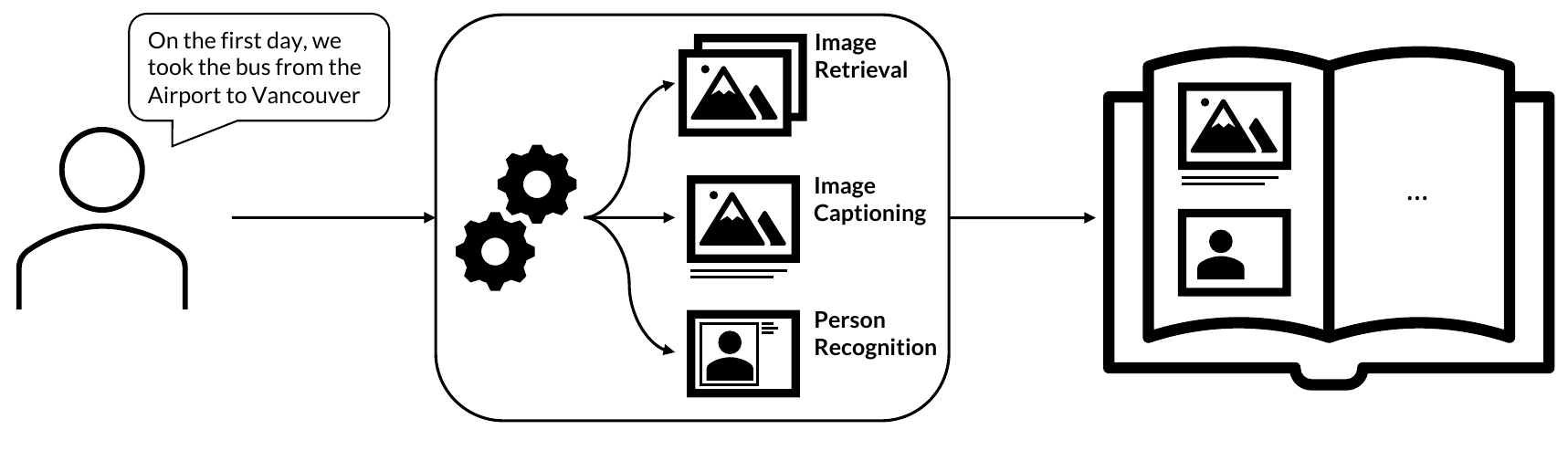}
    \caption{We plan to combine several modules based on deep learning models to create photo book pages from natural language input. These modules include, for instance, image retrieval, image captioning, and person recognition.}
    \label{fig:usecase_page_creation}
\end{figure}
\begin{figure}
    \centering
    \includegraphics[width=.85\textwidth,trim={0 10.25cm 9.3cm 0},clip]{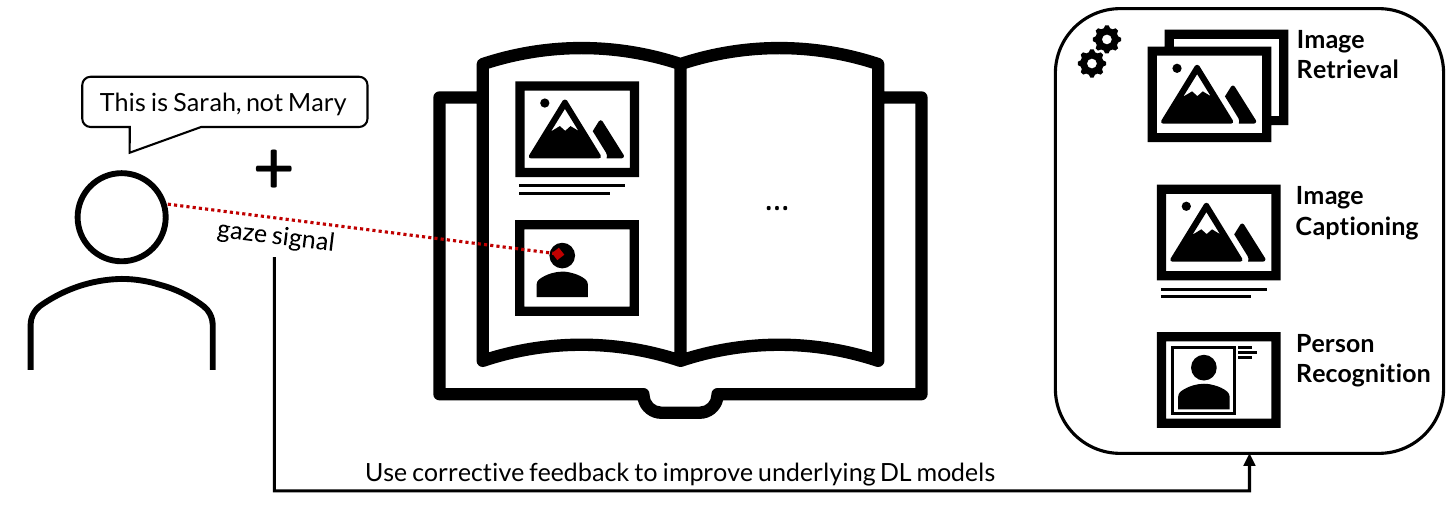}
    \caption{The user can provide multimodal feedback to the photo book tool to alter the created content. For instance, we plan to jointly interpret the user's gaze signal and spoken utterances to improve person recognition. An example is shown in figure \ref{fig:mmi_demo_setup}.}                           
    \label{fig:usecase_feedback}
\end{figure}

Thankfully, they find out about the AI software that integrates techniques developed within \ACRONYM{}. Using these, a photo book can be created by providing a set of images and by sequentially describing the occasion in natural language, be it a holiday trip or a wedding party. They can also describe the style and purpose of the photo book to guide the creation process. To make an example, imagine that they plan to create a photo book about their last family trip to Canada. They start off by telling the system: ``This will be a photo book for aunt Mary about our last trip to Canada. We would like to add some dramatic touch to it''. In return, the photo book creation tool suggests a suitable caption and basic style for the photo book. If not suitable, they can edit the caption or adapt the style, e.g., by selecting another frame type for captions or another font family. They would continue by describing how they perceived their vacation to the photo book tool just like they would describe it to another human: ``On the first day, we took the bus from the airport to Vancouver'' (see figure \ref{fig:usecase_page_creation}). As a response, the system creates a single page with suitable photos, i.e., from getting on the bus at the airport, a photo of the skyline of Vancouver from inside the bus and one with aunt Mary who was waiting for them at the bus stop. Since this is the first time family Smith is using this tool, the automatic caption generation module is uncertain whether its output is suitable and, hence, actively asks for feedback.

\begin{figure}
    \centering
    \includegraphics[width=\textwidth,trim={0 5.5cm 9.75cm 0},clip]{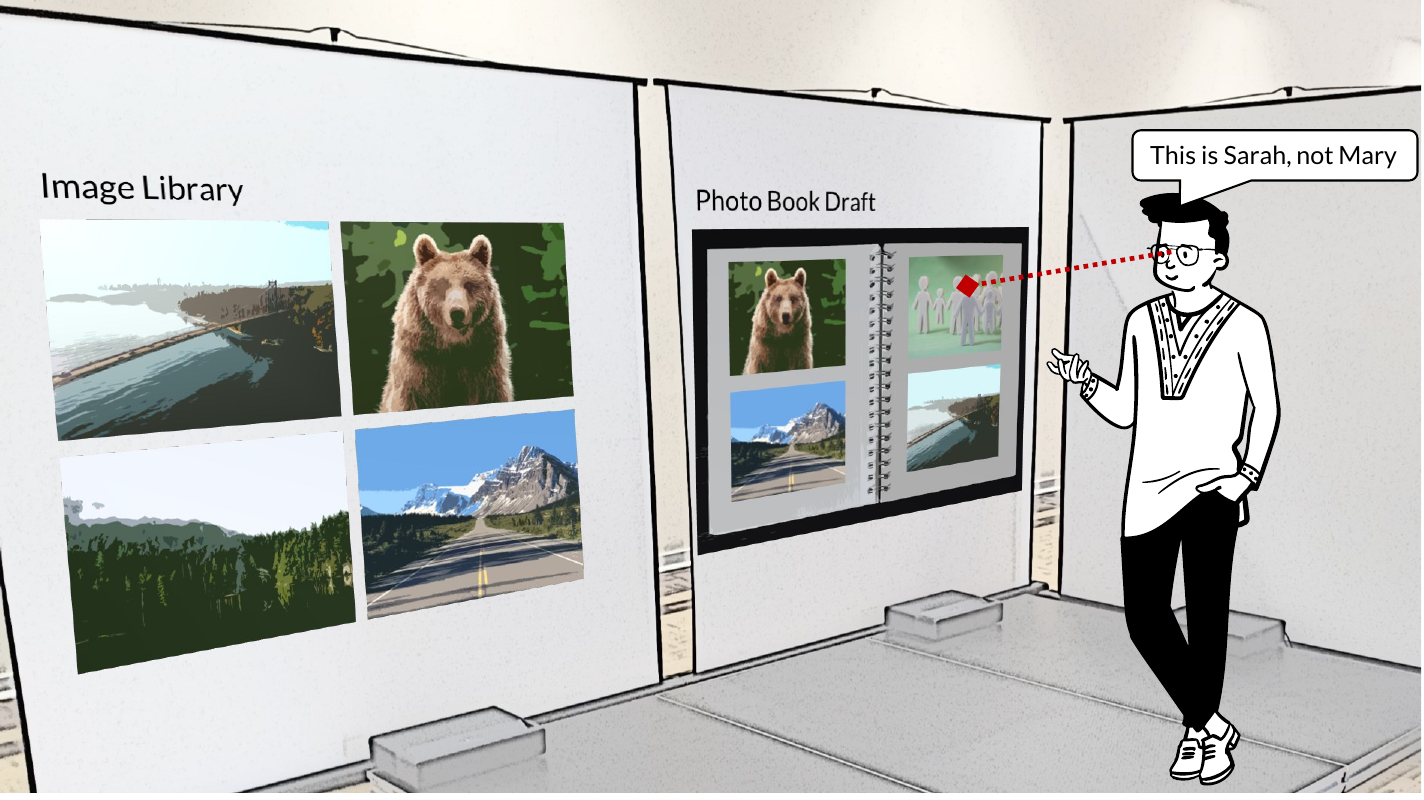}
    \caption{Example of a multimodal user input to our photo book application (based on an existing demo setup). The user provides corrective feedback in natural language by saying "This is Sarah, not Mary". The system uses his gaze to resolve the face that was referred to and uses the new information to update the underlying deep learning models as depicted in figure \ref{fig:usecase_feedback}.}
    \label{fig:mmi_demo_setup}
\end{figure}

\begin{figure}
    \centering
    \includegraphics[width=\textwidth,trim={0 1.5cm 0 2cm},clip]{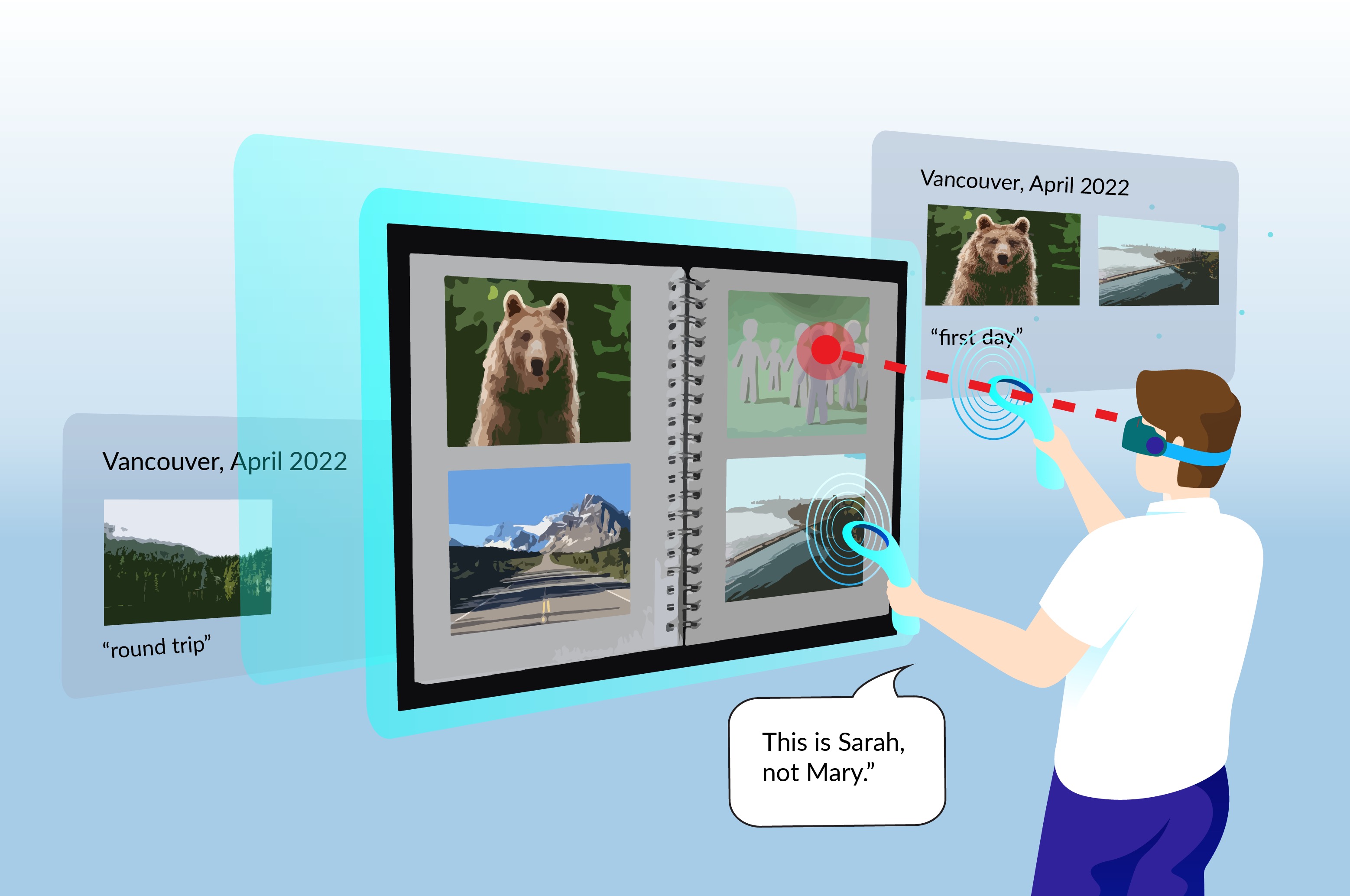}
    \caption{Visualisation of the virtual reality scenario. Images and the photo book are presented in an immersive virtual environment. Through multimodal interaction (pointing, eye-/gaze-tracking, natural speech) the user engages with the system and provides corrective feedback by saying "This is Sarah, not Mary". The system uses implicit and explicit pointing or gaze to resolve the face that was referred to and uses the new information to update the underlying deep learning models as depicted in figure \ref{fig:usecase_feedback}. In addition to the multimodal setup depicted in figure \ref{fig:mmi_demo_setup}, VR tracking provides detailed spatial tracking information that will be included in the data analysis.}
    \label{fig:vr_demo_setup}
\end{figure}

Being happy with this partial result, the family continues to describe the events saying ``The incident with the bears was extremely funny and the woods were so impressive''. The newly generated pages of the photo include pictures of the bear and the woods from their hiking trip, but none with aunt Mary, so they complain about this. ``Please add a picture with Mary here''. As the system does not know yet how Mary looks, it shows extracted faces from the provided photos and asks to select a picture of Mary. Mrs. Smith looks at a picture and says ``that's my sister Mary''. The system uses the gaze signal to identify the face that was referred to and learns to recognise Mary. Eventually, family Smith reports how their vacation ended: ``it was also something how aunt Mary had to take us to the airport on short notice because our car broke down and we almost thought we wouldn’t make it and how they welcomed us back at the airport after we landed.'' One of the images shows Sarah in front of aunt Mary's car, but the caption states ``This is aunt Mary after carrying us to the airport''. Mr. Smith corrects the system by saying ``this is Sarah, not Mary'' (see figures \ref{fig:usecase_feedback}, \ref{fig:mmi_demo_setup}, and \ref{fig:vr_demo_setup}). The system automatically corrects the caption and corrects the label for the detected face. From now on, the system will be better at differentiating between Sarah and her sister Mary. Alternatively, Mr. Smith could edit the caption to ``This is Sarah in front of her car after carrying us to the airport last minute.'' and the feedback contained in this post-edit would be used to update the image captioning model.

While the initial draft of the photo book is already quite good, the Smiths want to add some personal touch and they also spot some errors browsing through the suggestions. The system supports an immersive mode using \gls{vr} or just a normal desktop/tablet-based presentation. While they could use either mode and intuitive hand or touch gestures in combination with gaze-tracking/spoken dialogue to rearrange and edit each caption and photo by pointing or touching a photo and selecting from better alternatives presented by the system, by rating a photo as not suitable, or by providing feedback to a caption, the system also provides some higher-level tools: for story-based books the overall time and dramatic flow of the story and the included events are visualised along a time line (which works especially well in \gls{vr} thanks to almost unlimited virtual space). To avoid clutter, each event is represented by some iconic photos and a summarising caption, generated by the system. The Smiths can now put more or less emphasis on certain events, add or remove whole events, or ``zoom'' in and identify key characters and photos. For diary-type or location-centered books, the photos are clustered accordingly and visualised over a floating map and again the Smiths can now edit and provide feedback using rich multi-modal input. The system will continue to learn from the user input and actively ask for help in uncertain cases. The rich input/output modalities (especially in the \gls{vr} case) will benefit user and system on several levels. They will make active learning by the system more effective because multimodality can be used to disambiguate and to compensate for noise in single modalities. They will also improve the user experience because they allow for a more intuitive and effective interaction and visualisation and as they provide more data about the user to the system, the system can learn more effectively (using not only explicit but implicit inputs) about the user preference and can adapt the information load.


Over the past decade, researchers have studied similar scenarios~\cite{sandhaus_processes_2008} and proposed partial solutions for certain sub-task. For instance, different methods ranging from semantic modelling~\cite{sandhaus_semantic_2011} and meta data analysis~\cite{boll_metaxacontext-_2006,boll_semantics_2007} to deep learning solutions~\cite{withoft_ilmica_2022} have been investigated for retrieving and filtering photos according to general criteria or personal preferences~\cite{maszuhn_user-centered_2021}. Some of these works have also looked at data from social media activity to learn about user preferences or events~\cite{rabbath_automatic_2011,rabbath_multimedia_2011}. Other works have looked at the presentation layer, for instance, at how to create aesthetic layouts~\cite{sandhaus_employing_2011} or how to design novel augmented reality interaction techniques to allow users to easily annotate their photos~\cite{henze_whos_2011}. However, integrated solutions for a complete system are still missing, which highlights both the relevance but also the challenge of the presented scenario. While the goal of this project is not to develop a market-ready photo book application software, we are certain that we will be able to implement the use case as an AI testbed to extend the current state-of-the-art in interactive deep learning. We propose a unique and integrated approach that draws on our expertise from machine learning, NLP, multimodal interaction and HCI research. 

\section{Goals and Scientific Challenges of \ACRONYM}
\label{sec:Goals}
With the convergence of artificial intelligence and machine Learning, \gls{idl} is where the \gls{hci} community meets the \gls{dl} community \cite{sonntag2010ontologies,zacharias2018survey, amershi2014power,holzinger2016interactive,dudley_review_2018,teso2020challenges}.
\ACRONYM{}'s goals and scientific challenges centre around the desire to increase the reach of \gls{dl} solutions (and \gls{ml} solutions in general): \gls{dl} for non-experts in \gls{ml} and improving \gls{dl} models when not enough data is available (e.g., due to highly individualised tasks like photo book creation) or data quality is not sufficient.
In addition, to fully automate tasks in practical applications such as our use case of interactive photo book creation can be extremely difficult and even undesirable. As a consequence, our goals are to find a computational and design methodology to gracefully combine automated services with direct user input or manipulation.
We investigate our scientific goals in the context of our photo book application. However, the technologies developed shall be beneficial for other domains as well such as healthcare or smart manufacturing.
They can be summarised as follows: 

\begin{enumerate}
    \item Define and declare the role of humans in \gls{idl} (HCI): (1) realising the importance of studying users; (2) reducing the need for supervision by \gls{ml} practitioners; (3) explore interactivity in a tight coupling between the system and the user; (4) handle human ambiguity and confusion and instil trust and confidence through feedback and explanations; (5) explore gamification and serious games in the context of \gls{idl} and \gls{iml} in general.
    \item Provide a way for users to (1) understand why the system had made a particular prediction, and (2) adjust the (\gls{dl}) learner’s reasoning if its prediction was wrong. To this end, the system should provide an explanation for its predictions, and incorporate corrective feedback given by the user. How can this be done in practical terms? For providing useful explanations of model predictions, we will investigate the feasibility of solving tasks with interpretable (\gls{dl}) models rather than black box models \cite{RudinRa19}.
    \item Active and passive user input needs to be interpreted carefully to establish an efficient and effective interaction between humans and an AI system. The challenge includes to interpret signals from multiple input modalities (e.g., gaze and spoken instructions). It may be required to interpret the input signals according to a user or context model (e.g., reflecting a user's preferences or the interaction context). In \ACRONYM{}, we develop multimodal interaction techniques for incremental photo book creation with the goal to improve model training through rich multimodal user feedback and to improve the user experience through robust and intuitive interfaces. At the same time, we should avoid the limitations of human cognitive abilities.
    \item Implement mixed initiative interaction, an opportunity to explore interfaces that can leverage knowledge and capabilities of domain experts more efficiently and effectively. The \gls{ml} system and the domain expert should engage in a two-way dialogue to facilitate more accurate learning from less data compared to the classical approach of passively observing labelled data. In the context of our photo book use case, we aim at using, e.g., active learning and principles from human-in-the-loop expert systems. The greater goal is to perform application tasks more satisfactorily: human-machine teams shall surpass the efficiency/effectiveness of humans or machines in this task alone \cite{DBLP:conf/hhai/ZoelenMTFTCGBBN23}.
\end{enumerate}

\subsection{Natural Language Processing (NLP)} \label{sec:NLP}

Our approach for supporting photo book creation relies on several components based on 
deep learning models for image and multimedia data, in particular face and body shape recognition, text-to-image retrieval, image captioning, visual storytelling, and \gls{vqa} models. The different components are triggered based on a user's commands (e.g., "On the first
day, we took the bus from the airport to Vancouver" triggers the text-to-image retrieval component and the image captioning component). We plan to model this by either explicitly mapping triggering keywords to components, or by applying more sophisticated semantic parsers. The optimal way of processing user input will be determined in the course of the project based on insights from user studies.   
In this part of the \ACRONYM{} project, we investigate three core research problems associated with the application and interaction with \gls{dl} models in the context of our use case: (1) how to adapt state-of-the-art multimedia \gls{dl} models to process user-specific texts and images, which, in contrast to the generic data the models are usually applied to, requires to account for specific information related to the user and the events they want to present in their photo book; (2) how to improve the \gls{dl} components based on user feedback collected in the refinement phase based on the \gls{iml} paradigm; (3) how to use model explanations to achieve optimal interaction between user and model and best support the photo book creation process.

 Users have a personal relationship with objects and concepts displayed in the images of their photo book, and providing support in the photo book creation process requires modelling image content from a user's perspective. For example, we need to take into account that a user will refer to named entities in an image by proper name rather than a common noun (\emph{Mary} instead of \emph{a woman}). In \ACRONYM{}, we investigate how to adapt multimedia and multimodal \gls{dl} models to account for such user-specific information. For cross-modal (text-to-image) retrieval, we plan to implement state-of-the-art \gls{dl} retrieval models \cite{Zhang_2020_CVPR,jia2021scaling,Alikhani_Han_Ravi_Kapadia_Pavlovic_Stone_2022}, which retrieve items based on embedding similarities in a shared representation space, in combination with rule-based filters that take into account output from a person recognition model as well as available image metadata, such as time stamps and geolocation. For example, given a user query \emph{Show me the pictures of Peter and Mary playing football when we visited Vancouver}, the component retrieves images given the query \emph{Two people playing football} and returns the subset of images for which the person recognition model indicates Peter and Mary being present, and the geolocation indicates an image taken in Vancouver. In contrast to image captions that can be found in general purpose datasets such as MS COCO \cite{lin2014microsoft} or Flickr30k \cite{plummer2015flickr30k}, the captions generated by our captioning component should be (1) entity-aware (e.g., instead of generic descriptions of objects or concepts, the captions contain proper names for named entities), (2) stylised, and (3) controllable (see table \ref{t:image_captioning} for examples). Existing models for entity-aware captioning usually first generate a template caption with place-holders for named entities, which is then filled with information retrieved from associated text or knowledge bases \cite{lu-etal-2018-entity,biten2019good}. Ramnath et al. \cite{ramnath2014autocaption} propose an approach for personalised template-filling with information such as geolocation, time stamp, detected landmarks, recognised faces, which we plan to extend to incorporate finer-grained location information specified by the user. To generate stylised captions, we will explore caption generation reflecting sentiment \cite{Mathews_Xie_He_2016}, specific styles \cite{gan2017stylenet,guo2019mscap}, and taking into account a user's active vocabulary \cite{chunseong2017attend}. In the refinement phase, when additional captions are generated for newly retrieved images, the user should be able to exert fine-grained control over the concepts to be included in the caption, e.g., by actively modifying an abstract scene graph representation based on which the caption is generated \cite{chen2020say}. 
 In contrast to generating captions for images in isolation, the visual storytelling component generates a sequence of captions that form a coherent story for a retrieved sequence of images \cite{huang2016visual, Jung_Kim_Woo_Kim_Kim_Kweon_2020, wang2020storytelling}. Similar to the captioning component, the visual story component needs to be entity-aware and controllable. To this end, we will investigate to what extent approaches for adapting the captioning model can be transferred to the visual storytelling task. Finally, in the refinement phase, a \gls{vqa} component can directly answer the user's questions about image content, such as \emph{What was the name of the mountain in the background?}, or \emph{Did Peter join us for the trip to Lake Baikal?}. Here, we will focus on implementing models for answering questions that cannot be answered from information in the image alone, but require additional knowledge about named entities and specific events, that could for example be provided by a knowledge graph \cite{Shah_Mishra_Yadati_Talukdar_2019}.

In order to improve the above described components based on feedback collected in the photo book refinement phase, we implement an \gls{iml} framework that allows us to iteratively update the models based on new information via incremental and focused updates \cite{amershi2014power}. Training and improving the models in an \gls{iml} framework is crucial to our use case, as we cannot assume large amounts of labelled personalised data to be available at once, and therefore need to learn from user-specific data incrementally. 
In \ACRONYM{}, we explore how \gls{iml} can be applied to improve the multimodal \gls{dl} components for photo book creation, considering three scenarios: (1) debugging trained models, e.g., identifying and correcting spurious patterns learned by the model \cite{lertvittayakumjorn2021explanation}. Here, we assume an explanation-based interactive loop to be particularly helpful; (2) adapting pre-trained models to user-specific data with small amounts of annotations \cite{yao2021refining} (3) personalising models \cite{kulesza2015principles}, e.g., for generating captions following stylistic preferences of users. 
We focus on improving models based on explanatory feedback provided by the user, i.e., instead of providing only label-level feedback (e.g., a correct answer to a \gls{vqa} model), the user additionally provides information that states \emph{why} the provided answer is the correct one. Interacting on the basis of explanations has the potential to benefit both the user and the model: on the user side, providing richer feedback beyond the label level is in line with their preferred way of interaction \cite{amershi2014power,ghai2021explainable}. From the modelling perspective, learning from explanatory feedback instead of label-level feedback can improve data efficiency \cite{hancock2018training,ye2020teaching} and generalisation \cite{yao2021refining}.  We focus on the two most commonly considered types of human explanations, which are \emph{highlight explanations}, i.e., subsets of input elements deemed relevant for assigning a specific label; and \emph{free-text explanations}, i.e., natural language statements providing information about why specific label should be assigned \cite{wiegreffe2021teach}. Several ways for improving models (except for \cite{selvaraju2019taking} these were developed for models that process either text or image data) based on such human explanations have been proposed \cite{hase2021can,hartmann2021interaction}: using natural language explanations as additional inputs \cite{co-reyes2018metalearning,rupprecht2018guide,rajani2019explain}, using explanation generation as auxiliary task \cite{camburu2018snli,narang2020wt5,hase2020leakage,wiegreffe-etal-2021-measuring}, directly constraining intermediate representations \cite{selvaraju2019taking,ross2017right,shao2021right,rieger2020interpretations}, or exploiting explanations to generate additional training instances \cite{hancock2018training,Awasthi2020Learning,ye2020teaching,yao2021refining}. We will investigate how to combine and extend these methods to update multimodal \gls{dl} models based on multimodal feedback. Most of these approaches have only been tested in offline setups, where the model can be trained on the entire explanatory feedback at once. As a first step, we will investigate which methods are applicable in an interactive setup where models are updated incrementally. As all DL components process the same user-specific data, we assume that it might be useful to share user-specific information among the components by exploiting user feedback to update multiple components at once. To this end, we will experiment with a multi-task architecture with hard parameter sharing, which trains $n$ models for $n$ tasks with a subset of parameters being shared among them \cite{caruana1993multitask,collobert2011natural}, e.g., sharing the multi-modal encoder while maintaining task-specific classifier layers (or decoders for language generation tasks). By updating the encoder based on feedback collected for one task, the information will be available to models for the other tasks as well. For evaluating our methods for interactive deep learning, we will follow previous work in re-splitting existing task-specific datasets (e.g., Microsoft COCO \cite{lin2014microsoft} and Flickr30k \cite{plummer2015flickr30k} for image captioning and text-to-image retrieval, VQAv2 \cite{goyal2017making} and KB-VQA \cite{wang2017explicit} for visual question answering, VIST \cite{huang2016visual} for visual story telling) into new data splits that allow to evaluate specific model behaviour, e.g., if a model relies less on language bias \cite{agrawal2018don}, or if a model has better continual learning abilities \cite{del2020ratt,greco2019psycholinguistics}.

The central component of an \gls{iml} system is a tight interactive loop between user and \gls{ml} model, in which the model presents its current state of knowledge to the user, and the user provides feedback to the model accordingly \cite{amershi2011effective,dudley_review_2018,wang2021putting}. The former part of the loop could be supported by showing an explanation for why the model made a specific prediction or took a specific action. The ability to provide explanations for predictions, i.e., information about the reasons for why a specific prediction was made, is considered essential for large-scale adoption of AI systems by end-users \cite{gunning2017explainable,BARREDOARRIETA202082}. In \ACRONYM{}, we will investigate how to use model explanations to achieve optimal interaction between user and model. For \gls{dl} black-box models, this requires choosing an adequate mechanism to construct explicit representations of explanations that can be provided to the user \cite{kim2021multi}. While for image processing models, saliency methods can provide useful visualisations of important input regions, such methods are less intuitive for text inputs. Here, the compositional nature of language calls for more expressive attribution methods that can model interactions between input tokens \cite{bastings-filippova-2020-elephant}. We focus on the generation of suitable explanations for generative or predictive multi-modal tasks, e.g., by generating natural language explanations while at the same time marking image regions that were relevant for a prediction \cite{park2018multimodal}. For presenting the explanation to the target end-user, we investigate the personalisation of explanations \cite{tomsett2018interpretable,ras2018explanation,sokol2020explainability,ghai2021explainable,mohseni2021multidisciplinary} to elicit high quality feedback and increase user satisfaction. How to evaluate model explanations is an active research topic \cite{10.1162/tacl_a_00465,jacovi2020towards,deyoung2020eraser,doshi2017towards} and we will focus on using previously proposed metrics for comparing model-generated explanations with human-generated explanations on publicly available multi-modal datasets, in particular VQA-X and e-ViL \cite{vqax,kayser2021vil}.

The main deliverables of this part of the project are:
\begin{enumerate}
    \item  Implementation of multi-modal DL components for photo book creation support that are entity-aware and controllable. For image captioning and visual story telling, the components should be able to generate text in a specific style.
    \item Implementation of an IML framework which allows to update the DL components based on explanatory user feedback collected in the photo book refinement phase. In addition to learning from explanatory feedback, the model should retain its knowledge while learning new things, which calls for the application of continual learning methods \cite{d2019episodic,biesialska2020continual,Li2020Compositional} within the feedback loop to prevent catastrophic forgetting \cite{kirkpatrick2017overcoming}. 
    Incompleteness and uncertainty of human explanations \cite{tan2021diversity} should be accounted for when implementing a feedback mechanism into the model as a software package. To this end, we will build on insights from the core \gls{ml} part of the project that investigates the use of Bayesian modelling for feedback integration as described in section \ref{sec:ML}.
    \item  Implementation of XAI methods for multimodal models which provide explanations for black box \gls{dl} model  decisions and take into account user-specific information, e.g., background knowledge and the motivation for consuming the explanation.
\end{enumerate}

\subsection{Multimodal-Multisensor Interaction (MMI)}  \label{sec:MMI}

In No-IDLE, we aim at developing interactive training mechanisms that enable continuous improvements of \gls{dl} models. A central aspect of this interactive loop is human feedback. We investigate the effect of integrating multimodal user input on the effectiveness, efficiency, and usability of interactive model training. We target models of our photo book application which include, for instance, models for recognizing specific persons and objects (see section \ref{sec:ML}) and natural language generation models (see section \ref{sec:NLP}).

One goal is to implement a gaze-driven dialogue that can support the initial creation and iterative refinement of a photo book. The multimodal feedback from the user shall enable the underlying \gls{dl} models to learn new concepts, to differentiate between instances of a concept, and to improve the detection/recognition of know classes. We plan to implement simple state-based dialogues to realise interactive model training with human gaze as additional input modality (e.g., based on the open source dialogue platform Rasa\footnote{\url{https://rasa.com/open-source/}}). The goal is not to develop beyond state-of-the-art multimodal dialogue systems, but to investigate the effect of integrating gaze (or pointing gestures) in simple speech-based instructions on the usability and effectiveness of interactive machine learning systems. For instance, a face recognition model could wrongly detect Sarah as Mary as described in section \ref{sec:usecase}.
When the user detects that the person identification system failed, he could provide a corrective feedback in natural language: "This is Sarah, not Mary". The system should analyse the user's gaze to identify to which face he referred in his utterance. Figure \ref{fig:mmi_demo_setup} illustrates this interaction based on an existing demo setup with three wall-sized screens. 
This corrective feedback shall be used to improve the underlying deep learning models (see figure \ref{fig:usecase_feedback}).
While, in No-IDLE, we put a focus on gaze-based input, pointing gestures will be considered for this kind of reference resolution as well, especially in the context of AR/VR interaction settings or when interacting with a wall-sized screen. 
Also, multimodal interaction can benefit from system-initiated interaction. This is particularly interesting in combination with active learning techniques that shall be developed by the \gls{ml} group (see section \ref{sec:ML}).
We want to explore the effectiveness (does the system actually learn to recognise new persons and objects), efficiency (what time is required for the model until it can recognise a new class), and usability (is the system usable for lay users) of different approaches in collaboration with the HCI group (see section \ref{sec:HCI}).
Another goal is to produce captions that are more focused on what the user wants to describe. We  plan to integrate aggregated \cite{sugano2016seeing, cornia2018paying} or sequential \cite{takmaz2020generating,pont2020connecting,meng2021connecting} human attention traces estimated from the multimodal input signal (gaze and pointing) into the generation process.
We hypothesise that incorporating multimodal interaction signals can improve the robustness of and the user experience during the interaction with an interactive machine learning system. Eventually, this should improve the quality of human feedback and, hence, the efficiency of model updates during training. Also, we expect that multimodal interaction can lead to a better understanding of how a model works, to a better understanding of the model’s strengths and weaknesses, and eventually to more trust in the model’s decisions.

Human gaze is well known for carrying non-verbal cues that can be used intelligent user interfaces: the eye movement behaviour depends on the task in which a user is currently engaged \cite{deangelus_top-down_2009}, which provides an implicit insight into their intentions and allows an external observer or intelligent user interface to make predictions about the ongoing activity \cite{flanagan_action_2003,gredeback_eye_2015,rothkopf_task_2016,rotman_eye_2006}.
For instance, knowing which objects in a scene are fixated is a valuable context information for spoken feedback in personalised photo book creation. In particular, when deictic references must be resolved \cite{matuszek_grounded_2018,mehlmann_exploring_2014}. Also, there is a strong link between gaze behaviour and spoken language: speakers fixate elements ``less than a second before naming them'' \cite{griffin_what_2000} and the coordination of hand-movements depends on human vision, e.g., when ``directing the hand or object in the hand to a new location'' \cite{land_roles_1999}.
Human gaze can also be used to analyse or model the behaviour of a user (user modelling), e.g., to learn about a user's ongoing activity \cite{bulling_eyecontext:_2013,steil_discovery_2015}, their preferences \cite{lalle_gaze-driven_2021,barz_implicit_2022}, intentions \cite{huang_anticipatory_2016,barz_visual_2020}, or state \cite{Huang2019,bulling_cognition-aware_2014}.
Observing eye movement behaviour during interaction with an interactive machine learning system could reveal situations in which the user disagrees with the model output. If these situations coincide with the model being uncertain about the output, this may be a good point in time to trigger a feedback request to the user (system-initiative).

In \ACRONYM{}, we focus on human gaze and pointing gestures as additional interaction modalities. We investigate the impact of using multimodal interaction signals on recognising objects or persons as context-information and to personalise the natural language generation process in the context of the photo book creation and refinement process. The challenge is that relevant persons and objects, their appearance, or similar properties can significantly vary between users and the occasion for creating such a book \cite{barz_automatic_2021_sensors}.
However, pre-trained models cannot account for such dynamic circumstances and adaptive models or agents are required that incrementally and continuously learn from human collaborators or interlocutors. 
The main deliverable is a software extension of an existing DFKI system, the multisensor-pipeline (MSP)\footnote{\url{https://github.com/DFKI-Interactive-Machine-Learning/multisensor-pipeline}}. The resulting modules shall be integrated and evaluated in the photo book creation process based on the experimental procedure as depicted in section \ref{sec:hci_evaluation}:
\begin{enumerate}
    \item Implementation of a module that enables to learn about unseen classes (objects) when the context shifts (class-incremental learning) and to improve the recognition of known classes via multimodal user interaction based on, e.g., transfer learning \cite{kading_fine-tuning_2017} and active learning (see section \ref{sec:ML}).
    Similarly, we plan the implementation of a module to differentiate between multiple instances of the same class trough multimodal human-machine interaction. We focus on the differentiation between multiple persons according to our photo book use case (e.g., to filter for images showing a particular person). This part will benefit from novel active learning approaches as described in section \ref{sec:ML}. We will also investigate in how far this module can be used to track meta information like ownership (see the COPDA project\footnote{\url{https://www.dfki.de/en/web/research/projects-and-publications/projects-overview/project/copda}}). Real-time tracking of multiple instances could be achieved by a combination of (multi-)object tracking \cite{li_siamrpn_2019} and models that estimate object properties such as colour, size, and shape \cite{thomason_learning_2016}. Such models are of particular interest when grounded in natural language, which would facilitate expressive explanations for classification results (related to section \ref{sec:NLP} and the XAINES project\footnote{\url{https://www.dfki.de/en/web/research/projects-and-publications/projects-overview/project/xaines}}).
    \item Implementation of a module of a new image clustering and object tracking method that can help "quick start" multimodal interactive model training upon domain shifts, because a single (user-provided) label can be propagated to multiple samples, e.g., to an image cluster or to samples from object tracking (semi-supervised learning). The idea is to cluster fixated image contents in the photo book application and, once a label is provided via speech, to propagate this label to the whole cluster. Similarly, \gls{fsl} from the ML task (see section \ref{sec:ML}) should help to overcome this cold-start problem. Few-shot image classification \cite{wertheimer_few-shot_2021} or few-shot object detection \cite{fan_few-shot_2019} enables image classification or object detection, respectively, using around five example images.
    \item Implementation of multimodal interaction techniques based on eye tracking for the photo book application. This includes approaches to provide feedback on model outputs multimodally (e.g., correcting labels for misclassified persons, triggering post-editing of generated captions, and guiding the caption generation process), but also general multimodal interaction with photo book representations in desktop or VR settings (for instance rearranging images, selecting better photos, or similar selection and manipulation actions).
\end{enumerate}


\subsection{Machine Learning (\gls{ml})} \label{sec:ML}

An important factor which contributes to the recent success of \gls{dl} (apart from superior computing power and training algorithms) is the availability of labelled data. 
In fact, neural networks are known to be data-hungry (e.g., popular benchmark datasets range from tens of thousands of labelled samples as in CIFAR-10 to millions as in ImageNet dataset). 
However, data labelling is a costly, human labour intensive activity. In certain domains such as healthcare and biomedicine where considerable expertise may be required, data labelling becomes a limiting step in the realisation of the value of \gls{ml}.
This is also the case for the creation of personalised photo books. For instance, when the system should learn to differentiate between faces and body shapes of a set of persons in order to select images containing them or not while the persons may differ per user and photo book.
Thus, it is imperative to build \gls{ml} algorithms which are capable of learning from significantly fewer labelled samples to save human time. 

A set of methods known as active learning \cite{Settles2010, monarch2021human} tackle this problem by allowing the system to identify a subset of maximally informative samples from a given pool of unlabelled data to be queried for additional labelling/feedback.
In the context of \gls{iml} in this proposal, active learning plays a key role in how a learning system requests, receives, and learns from user input.
In combination with the \gls{hci} tasks (section~\ref{sec:HCI}), this forms a joint task for mixed-initiative interaction: \gls{ml} system and human domain expert engage in a two-way dialogue, facilitating learning from less data compared to the classical approach of passive consumption of labelled data.
One direction to explore are new input techniques that allow users to provide more informative feedback \cite{ratner2016data}, compared to traditional low dimensional labels.

Popular methods in active learning might be uncertainty-based \cite{Konyushkova2019, Tong2001, Joshi2009}, density- or diversity-based approaches \cite{Gissin2019,Sourati2018}, ensemble methods \cite{Beluch2018, McCallumzy1998, Freund1997},  and expected error reduction \cite{Roy2001}.
A common problem of pure uncertainty-based methods is that the selection strategy depends on the performance of an existing model.
This could be problematic in the early phase of training since outcomes are likely to be unreliable, leading the algorithm to query poor examples and thus lead to inefficiencies.
Similarly, in pure density-based approaches data labelling could be redundant if the present model produces already high confident predictions.
Recently, methods have been proposed which try to mitigate this problem by combining and balancing uncertainty and diversity of the new samples w.r.t.\ the data distribution \cite{Ash2020,Smailagic2018,Ozdemir2018,Yang2017,Huang2010}.
Bayesian approaches have also been proposed \cite{Kirsch2019, Gal2017, Kapoor2007}, but they do not scale well to deep networks with large datasets. Other recent works include Fisher information \cite{Ash2021,Sourati2018} and learning to select from data \cite{Konyushkova2017}.

We tailor active learning technologies to be applied in No-IDLE in the context of our photo book scenario. The goal is to train a model that is able to differentiate between individual persons contained in a set of photos with little labelling effort by the user. The basis for this feature are computer vision models that enable a robust detection and location of faces and body shapes. For any set of images, these models can provide a pool of unlabelled face and body images. This is helpful to filter for images showing humans versus, e.g., landscape photos. However, for personalised photo books, we want the system to be able to differentiate between individual persons to filter for photos with specific persons. For instance, a user request could be “please add an image of Mary in front of our rental car”. The persons involved may vary as they are highly dependent on the user and the occasion for creating the photo book.
We will (1) implement and evaluate new sampling techniques/active learning approaches that enable model training with small amounts of labelled data and (2) investigate when system-initiative feedback requests should be shown and how they should be designed in order to maintain a good user experience. A good opportunity to trigger a feedback request could be right after a user takes the initiative to provide a new name (i.e., a label) for a person/face or corrects a label. For instance, if a user tells the system "this is Mary", the system could query for the most informative unlabelled instances that may also show Mary like "Ah, this is Mary. I guess, I've seen her on other pictures too. Is this Mary again [system shows another face image]?".

In this proposal, we aim to address the following \gls{ml} problems:
\begin{enumerate}
    \item On the experimental side, we first investigate the performance of existing uncertainty functions for various neural network architectures on image classification/segmentation tasks (see, e.g., figure \ref{fig:retina}).

    \item On the experimental side, this point is related to studying if we should only use the DL black box models in the IML process when we perhaps do not need to. The point brought forward in \cite{RudinRa19} is that one might consider that (in IML) maybe interpretable deep-learning models can be constructed, or transparent models be used in conjunction with DL models according to the user feedback. In machine learning, these black box models are created directly from data by an algorithm, meaning that humans, even those who design them, cannot understand how variables are being combined to make predictions. Also see surrogate models for this purpose, co-creating a transparent model from the predictions. A global surrogate model is an interpretable model that is trained to approximate the predictions of a black box model. We can draw conclusions about the DL black box model by interpreting the surrogate model \cite{DBLP:journals/jair/BurkartH21}. 
    
    \item On the practical side including Few-Shot-Learning: the motivation for this \gls{ml} task comes from MMI (section \ref{sec:MMI}),
    where we want the system to learn new objects during an interactive training session with the user, given that the user has provided feedback/labels for a few examples.
    In the literature, this problem could be tackled using techniques from \gls{fsl} \cite{Tian2020}. 
    The main challenge is how to learn a good latent embeddings of the inputs and the labels, and to align them together in such a way that certain attributes from both inputs and labels can be transferred to unseen objects.
\end{enumerate}

The research outcomes and main deliverables will include the design of new uncertainty functions, which will be used in \gls{iml}-related tasks such as \gls{nlp} (section \ref{sec:NLP}) and \gls{mmi} (section \ref{sec:MMI}).
Additionally, together with \gls{hci} (section~\ref{sec:HCI}) we will promote an active role for the human-in-the-loop:
besides providing labels, we want to explore different ways of providing/correcting explanations, aligning important features learned by the machine with human intuition, interpreting learned models, and finding a common ground with general \gls{hci} tasks, including a more generic approach for generating explanations and insights into the effectiveness of few-shot learning.

\subsection{Human-Computer Interaction (HCI)} \label{sec:HCI}

We explore the role of humans in \gls{idl}. From our own previous work~\cite{herrlich_instrument-mounted_2017} and from the literature~\cite{picard_affective_2000,oviatt_human-centered_2006, ryan_self-determination_2000}, the relevance of motivation, emotion and factors like cognitive load on how interfaces and systems are used and, consequently, how these factors should be taken into account during interface design is quite clear. \gls{idl} presents both a potential solution and an additional challenge in this regard~\cite{amershi2014power}. Furthermore, we want to transfer insights from our previous works in the medical domain and virtual reality. We have studied expert users such as medical doctors\footnote{\url{https://medicalcps.dfki.de/wp-content/uploads/2017/08/KDI_V2_Pro_v04_2.mp4}} and explored VR for \gls{idl}, e.g., for image classification in VR~\cite{prange2021demonstrator}, and as a general prototyping and evaluation environment for human-centered interaction design~\cite{reinschluessel_virtual_2017,klonig_integrating_2020,vera_eymann_quantifying_2021,omar_jubran_expanding_2021,queck_spiderclip_2022}. 

Referring to the example ``photo book'' application scenario described above, we plan to explore the combination of VR and \gls{idl} as a multi-modal, immersive interaction environment. This environment supports rich data input signals, for example, gaze and eye tracking, tracking of spatial movements and features such as pointing using a controller or freehand gestures and recording 3D trajectories over time as well as audio and speech input. It also integrates multi-modal output signals in the form of 3D graphics, spatial audio and simple forms of tactile feedback. Last but not least, it provides unlimited virtual space. 
As we sketched in the application scenario, we want to investigate how to leverage the potential of VR for \gls{idl} but the VR environment also provides an ideal test bed for generating and comparing data and models to be used in the real world because it is much easier to control and deploy. While existing works in this area have investigated specific components and tasks of the example usage scenario, e.g., the selection of aesthetically pleasing photos~\cite{withoft_ilmica_2022}, taking a specific look at the human factors with respect to the rich input and output modalities within virtual reality is a novel idea and has not been explored in the context of \gls{idl} to the best of our knowledge.

From an \gls{hci} perspective, the goals can be summarised as exploring new ways for learning systems to interact with their users, namely:  
(1) how user-driven learning cycles can involve more rapid, focused, and incremental model updates; 
(2) how to reduce the need for supervision by \gls{ml} practitioners; 
(3) As a result of these rapid interaction cycles common in \gls{iml}, even users with little or no machine-learning expertise should be able to steer machine-learning behaviours through low-cost trial and error or focused experimentation with inputs and outputs. How can this be supported from the \gls{hci} perspective? 
(4) Transparency can help provide better labels (contextual features, \gls{ml} predictions, etc.) towards explainable \gls{iml}. The experimental setup should include explainable \gls{iml}, where the user feedback is derived after the system explains its results, to avoid ``right answers for the wrong reasons'', see, e.g., \cite{anders2022finding}.   
(5) Understanding how people actually interact---and want to interact---with machine-learning systems is critical to designing systems that people can use effectively \cite{DBLP:journals/corr/SimardACPGMRSVW17}. 

More specifically, we plan to study basic properties like mental and physical load, attention split problems, confusion, and emotional affect. These provide the foundation to investigate more complex effects regarding user intention and strategies, trust, and confidence in using the system.
Furthermore, we expect a large impact of explainability techniques on these factors. We plan to experiment with different graphical and textual or spoken explanations. By studying these factors from the user's perspective we intend to optimise the effectiveness of active learning techniques.

We plan to run comparative studies within VR, for example, exploring different interaction designs, information presentation and \gls{dl} techniques. The idea is to measure human factors as listed above, e.g., cognitive load, but also other factors of the user experience, such as emotional affect and motivational measures such as user engagement and study their impact on active learning efficiency and effectiveness.

Considering the potential effect of user motivation, experimenting with forms of gamification~\cite{deterding_game_2011,deterding_gamification_2011} and serious games within the framework of the example scenario seems relevant. One approach in that regard will be to turn the respective task, e.g., finding photos with certain contents, describing a picture, sorting or clustering pictures, inserting a missing or best fitting picture into visual photo book story, into challenges by introducing a time limit (soft or hard), rewards (short, mid, long term) and potentially forms of social relatedness (synchronous or asynchronous forms of multi-player). Gamification could also be used to provide a measurement of the quality of the \gls{dl} model by using it to acquire user ratings of the overall output.

As a side note, to facilitate user participation in our experiments, we plan to set up an open lab space in the centre of the city of Oldenburg (in the CORE Oldenburg) to increase participation and recruit volunteers with diverse demographic backgrounds.

The main deliverables in this area are: 
\begin{enumerate}
    \item Implementation of different interaction modalities within virtual reality, e.g., free hand gestures vs. controller based selection or manipulation vs. \gls{nlp} and possible combinations.
    \item Studies about the influence of conscious and unconscious gestures, e.g., certain movements or posture that relate to confusion or decision insecurity; gaze or eye tracking (here there is a very strong link to multimodality).
    \item Implementation of different feedback forms and modalities to encode information about the \gls{dl} results and decision process, from “simple” visual features (colour, location, etc.) to audio or tactile channels.
    \item Concepts and studies of the effect of more playful approaches (serious games and gamification) with respect to user motivation and user feedback quality and quantity for \gls{idl}.
\end{enumerate}

\subsection{Evaluation Plan} \label{sec:hci_evaluation}

In this subsection we provide details about our general evaluation process and study plan. Of course, due to the novelty of the research, the plan will have to be adjusted throughout the project as it depends on the progress and results of the technical parts and work packages. We want to emphasise that the guiding overall focus of all evaluation activity is to investigate and improve the \gls{idl} process as discussed in the specific subsections, e.g., how can the observed user behaviour and user experience be utilised as a means for improving efficiency and effectiveness of \gls{idl}. This also is reflected in the way that VR is used within this project, i.e., as powerful tool for studying user behaviour and collecting data using photo book creation as an example application as opposed to investigating the use of VR for photo book creation, which is explicitly not a focus point of this project.

Firstly, we plan to conduct a number of smaller studies that look at very specific aspects and that lay the foundation for a larger study towards the end of the project. At the beginning, we will focus on fundamentals and isolated elements and shift to investigating more complex combinations of system features and tasks over time. This will also be reflected in the methods we apply. At the beginning we will employ methods of a more exploratory and formative type, for instance, case studies using methods such as interviews, cognitive walk-troughs, think-aloud, observation and forms of moderated discussion. Of course, this does not exclude also collecting quantitative data already in this phase if possible.

The main study approach of a more summative character will be using an experimental setup comparing two conditions (control + intervention) or (if applicable) a factorial design with up to three or four conditions using appropriate tools and collecting quantitative measures like completion times, labelling accuracy in addition to (preferably validated) questionnaires for subjective feedback especially for measuring user experience and usability, e.g., SUS~\cite{brooke_system_1986}, PANAS-X~\cite{watson_panas-x_1994} and other SDT-based~\cite{ryan_self-determination_2000} tools related to motivation and also physical and mental load (e.g., NASA-TLX~\cite{hart_development_1988,hart_nasa-task_2006}).

The final decision for the experimental design with respect to independent or dependent groups (within-subjects vs. between subjects design) hinges on factors like the expected learning effect vs. fatigue effects and is subject to the specific experimental design for each study based on testing and pre-studies to quantify these confounding effects.

In addition, the VR setup in particular but also the eye-tracking scenario provide unique opportunities to collect objective data, most importantly, eye-tracking and movement data, e.g., trajectories of the controllers. We will also look into additional psycho-physiological measures, such as heart rate that are relatively easy to measure with off-the-shelf wearables.

We will base the number of participants on comparable studies and standards in HCI, typically in the range of 20-80 participants per experiment. The general experimental procedure includes the following steps:
\begin{enumerate}
    \item Introduction and welcome of participants and collecting their informed consent.
    \item A training or accommodation phase, which is especially important in the VR case.
    \item A calibration phase or procedure, which can also include collecting base levels of certain measures.
    \item The main part, i.e., participants perform specified tasks under different conditions, e.g., different forms of visual feedback, input gestures or active learning prompts. Some data are collected continuously through logging other data (e.g., subjective feedback) are collected after each condition (in accordance to the respective measure or questionnaire).
    \item Collection of post-experimental and independent data (e.g., demographics).
    \item De-briefing and ``Goodbye''.
\end{enumerate}

Throughout the procedure participants will be able to take breaks as needed (especially in the VR scenario) and we will adhere to scientific standards including getting approval of the DFKI ethics committee. The statistical analysis of individual measures will be carried out using linear models such as ANOVA for comparing means or non-parametric tests like Friedman~\cite{cairns_doing_2019}. In addition, forms of time series analysis and clustering will be looked into for analysing and correlating spatial measures such as body, hand, or controller movements.
We will also consider post-hoc experiments based on recorded user inputs to test additional \gls{iml} approaches. This can be done by simulating the interaction signals of our study participants if the model outputs have no immediate impact on the interaction flow.

\section{Existing Hardware and Software Frameworks at DFKI IML}


By harnessing the power of foundation models \cite{Ali2019}, i.e., any \gls{ml} model which is trained on a large-scale dataset and can be adapted to a wide range of downstream tasks, the research community is optimistic about their social applicability \cite{Bommasani2021}, especially in the healthcare discipline with integrated human interaction. Especially, patient care via disease treatment usually requires expert knowledge that is limited and expensive. Foundation models trained on the abundance of data across many modalities (e.g., images, text, molecules) present clear opportunities to transfer knowledge learned from related domains to a specific domain and further improve efficiency in the adaptation step by reducing the cost of expert time. As a result, a fast prototype application can be employed without collecting significant amounts of data and training large models from scratch. In the opposite direction, end-users who will directly use or be influenced by these applications can provide feedback to power these foundation models toward creating tailored models for the desired goal of \gls{idl}, based on DFKI IML's existing software frameworks: \cite{zacharias2018survey,Nunanni2021,Nguyen2020,sonntag2020skincare}.

The planned multimodal multisensor interfaces in \ACRONYM{} will be based on the multisensor-pipeline (MSP)\footnote{\url{https://github.com/DFKI-Interactive-Machine-Learning/multisensor-pipeline}}, our lightweight, flexible, and extensible framework for prototyping \gls{mmi} based on real-time sensor input \cite{barz_multisensor-pipeline_2021}. The MSP ecosystem will benefit from the developments in \ACRONYM{}, because novel modules will be released as open source to the research community.
\ACRONYM{} will take advantage from recent and upcoming developments in the BMBF Project GeAR\footnote{\url{https://www.dfki.de/en/web/research/projects-and-publications/projects-overview/project/gear}} (ends in September 2022): we are developing methods that reduce the human effort in the process of annotating mobile eye tracking data as described in \cite{barz_automatic_2021_sensors}. In GeAR, we target semi-automatic annotation for analytical applications (post-hoc) rather than real-time interactive model training, which is integrated into the application itself.

\section{Existing Application Domains and Demo Scenarios at DFKI IML}
\label{applicationdomains}

\begin{figure}
	\centering
    \includegraphics[width=.85\textwidth,trim={1cm 0 0 1cm},clip]{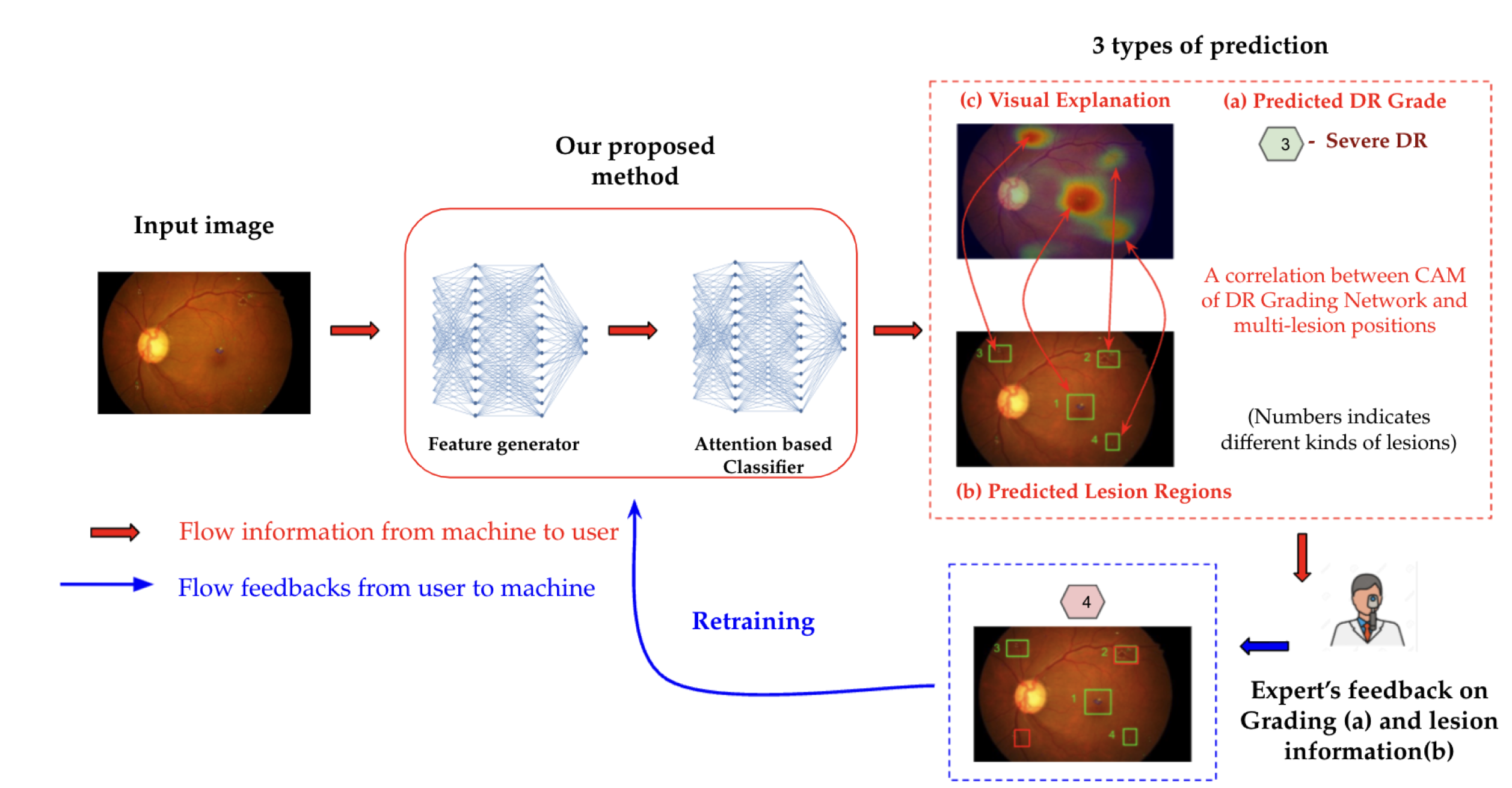}
    \caption{High level overview of our proposed method in the \gls{idl} workflow of the Ophthalmo-AI project (BMBF). Given a retinal image, our \gls{dl} models will generate 3 types of predictions (DR grade, lesion region, visual explanation) simultaneously. Ophthalmologists can observe the predictions and provide feedback for model fine-tuning.}~\label{fig:retina}
\end{figure}

\begin{figure}
	\centering
    \includegraphics[width=.85\textwidth,trim={0.8cm 9.4cm 6cm 0}, clip]{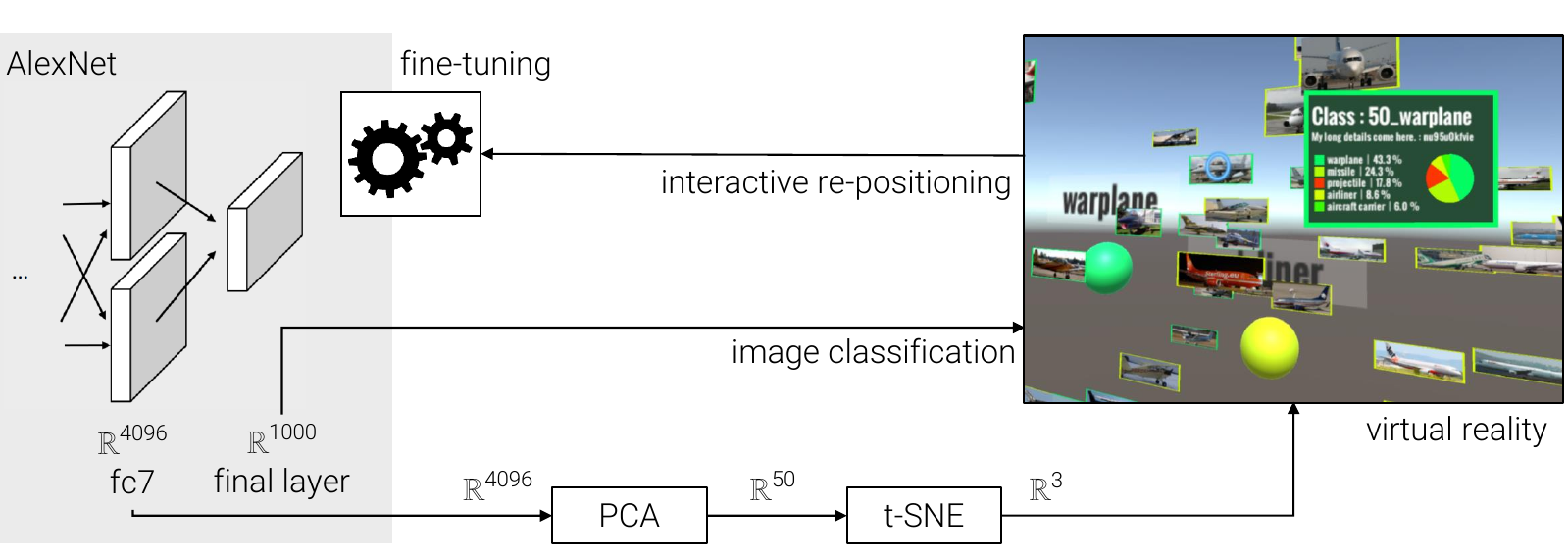}
    \caption{Architecture of our approach in \cite{prange2021demonstrator} based on PCA and t-SNE dimensionality reduction. Based on a pre-trained AlexNet we calculate 3D coordinates for each image. In \gls{vr}, information related to a particular image is displayed if the user looks at it. 
  	}~\label{fig:architecture}
\end{figure}

\begin{figure}
	\centering
  	\includegraphics[width=.85\textwidth,trim={0 .2cm 0 0},clip]{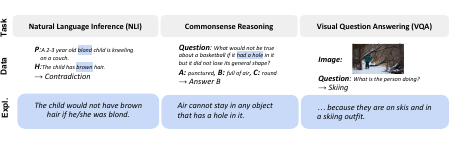}
\caption{Examples of existing datasets with human explanations for natural language inference \cite{camburu2018snli}, commonsense reasoning \cite{rajani2019explain}, and visual question answering \cite{park2018multimodal}. Explanations are either free-form (bottom line) or subsets of the input data (highlights in blue). These datasets can be used for both learning to generate natural language explanations as well as simulating explanatory feedback fed to the model in the sense of explanatory \gls{iml}, see \cite{teso2019explanatory}, where in each human-in-the-loop step, the learner explains its prediction to the user, and the user can provide explanatory feedback back to the model in order to improve it. Whereas explanatory IML mainly focuses on correcting \textit{right for the wrong reason} behaviour, we we will also explore how to use explanatory feedback to adapt models to user-specific input data.}~\label{fig:explanatory_feedback}
\end{figure}

\begin{figure}
	\centering
  	\includegraphics[width=.95\textwidth]{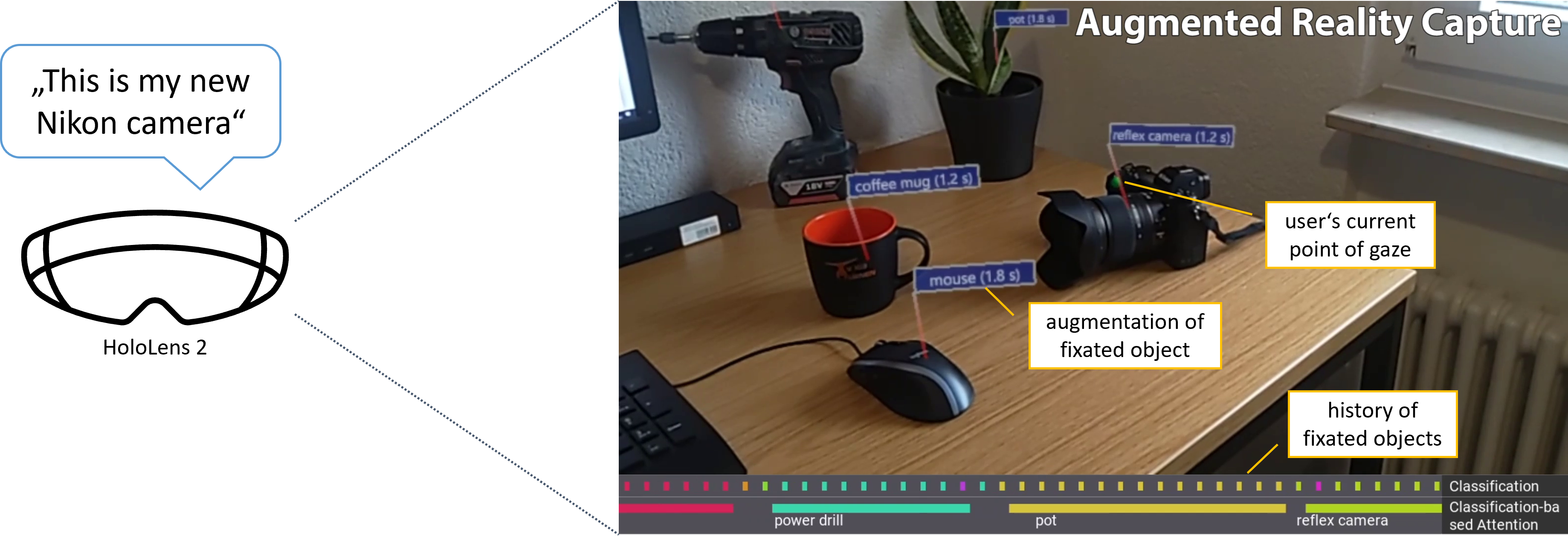}\vspace{.2cm}
\caption{Our prototype based on Microsoft's HoloLens 2 classifies and augments fixated objects in real-time \cite{barz_automatic_2021_etra}. It displays classification labels and the duration of recent attention events to the user as a hologram. The demo video can be viewed here: \url{https://www.youtube.com/watch?v=bdNClVz9ylE}. In \ACRONYM{}, we plan to enable interactive model adaptation based on foundation models: For instance, the user could create a specific instance of "reflex camera" and name it "Nikon camera" via speech (as shown in the image). This is related the COPDA project which aims to establish and maintain object relations like ownership. Other examples include that users may correct wrong classifications or teach new classes to the \gls{ml} system in a mixed-initiative dialogue. }~\label{fig:gaze}
\end{figure}

We build the MMI and HCI components of this project upon four past application domains and demo scenarios, which we detail in the respective figure captions:

\begin{itemize}
    \item Interactive Doctor Feedback (use case from BMBF Ophthalmo-AI\footnote{\url{https://www.dfki.de/en/web/research/projects-and-publications/projects-overview/project/ophthalmo-ai}}) project (see figure \ref{fig:retina})
    \item Interactive Image Classification in \gls{vr} (see figure \ref{fig:architecture})
    \item Explanatory \gls{iml} (use case from XAINES project, see figure \ref{fig:explanatory_feedback}): In XAINES, we develop models that provide explanations for predictions in an explanation-feedback loop,  which can serve to improve the model based on human feedback, and to personalize explanations. These models will serve as a starting point for developing interactive DL models for the \ACRONYM{ }photo book use case. 
    \item The multimodal interaction systems in \ACRONYM{} will be build based on our experience and outcomes from recent research projects (SciBot, GeAR). This includes methods for real-time interpretation of multimodal sensor streams such as mobile eye tracking data \cite{barz_visual_2020,barz_automatic_2021_etra,barz_automatic_2021_sensors,barz_implicit_2022,bhatti_eyelogin_2021,kapp_arett_2021} (for an example, see figure \ref{fig:gaze}), but also pen-based input signals \cite{barz_digital_2020}. In addition, we will use and further develop our framework for building multimodal, real-time interactive interfaces, the \emph{multisensor-pipeline} \cite{barz_multisensor-pipeline_2021}.
\end{itemize}

\vspace{-.25cm}

\begin{table}
\centering
\renewcommand{\arraystretch}{2}
\begin{tabular}{cll}
 \multirow{4}{*}{\includegraphics[width=.35\textwidth]{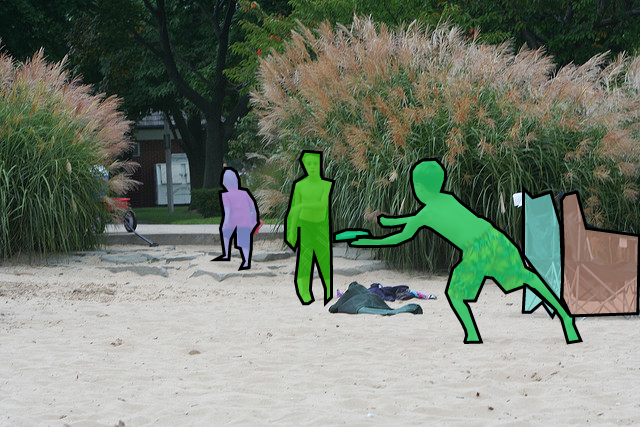}} & Generic: & Two boys are playing frisbee on the beach. \\
 & Personalised: & Peter and Tom are playing frisbee at the Pyla campsite. \bigstrut\\
& Stylised: & A heated game of frisbee on the Pyla court. \bigstrut\\
& Controllable: & David has enough and returns to the cabin. \bigstrut\\
\end{tabular}
\caption{Example image captions for an image taken from MS COCO, showing the difference between generic image captions and entity-aware, stylised, and controllable image captions require for photo book creation support.}
\label{t:image_captioning}
\end{table}

\section{Conclusion}
We presented the anatomy of the No-IDLE prototype system (funded by the German Federal Ministry of Education and Research) and described  basic and fundamental research in interactive machine learning while addressing users’ behaviours, needs, and goals. We decribed goals and scienfific challenges that centre around the desire to increase the reach of interactive deep learning solutions for non-experts in machine learning, followed by a methodology for interactive machine learning combined with multimodal interaction which will become central when we start interacting with semi-intelligent machines in the upcoming area of neural networks and large language models. Future work includes "No-IDLE meets ChatGPT". The overall objective of this follow-up project will be to leverage the opportunities arising from large language models and technologies for the No-IDLE project. No-IDLE aims to enhance the interaction between humans and machines for the purpose of updating deep learning models, integrating cutting-edge human-computer interaction techniques and advanced deep learning approaches. Considering the recent advances in LLMs and their multimodal capabilities, the overall objective of "No-IDLE meets ChatGPT" should be well motivated.

\begin{acks}
    This work is funded by the \grantsponsor{bmbf}{German Federal Ministry of Education and Research}{https://www.bmbf.de/} under grant number \grantnum{bmbf}{01IW23002}.
\end{acks}

\bibliographystyle{ACM-Reference-Format}
\bibliography{references}



\end{document}